%% file: main.tex
\documentclass[10pt,twocolumn,letterpaper]{article}

\usepackage{iccv}
\usepackage{times}
\usepackage{epsfig}

\usepackage{graphicx}
\usepackage{amsmath}
\usepackage{amssymb}
\usepackage{booktabs}
\usepackage{multirow}
\usepackage{subcaption}

\input{math_commands/math_commands.tex}

\usepackage[pagebackref=true,breaklinks=true,letterpaper=true,colorlinks,bookmarks=false]{hyperref}
\usepackage[capitalize]{cleveref}
\crefname{section}{Sec.}{Secs.}
\Crefname{section}{Section}{Sections}
\Crefname{table}{Table}{Tables}
\crefname{table}{Tab.}{Tabs.}

\iccvfinalcopy 

\def\iccvPaperID{5371} 

\begin{document}

\title{HyperStyle3D: Text-Guided 3D Portrait Stylization via Hypernetworks}

\author{Zhuo Chen$^{1}$ \quad 
Xudong Xu$^{2}$ \quad 
Yichao Yan$^{1}$\thanks{Corresponding author} \quad 
Ye Pan$^{1}$ \quad
Wenhan Zhu$^{1}$ \quad \\
Wayne Wu$^{2}$ \quad 
Bo Dai$^{2}$ \quad 
Xiaokang Yang$^{1}$ \\ 
$^1$ MoE Key Lab of Artificial Intelligence, AI Institute, Shanghai Jiao Tong University \\
$^2$ Shanghai AI Laboratory\\
{\tt\small ningci5252@sjtu.edu.cn xuxudong@pjlab.org.cn }\\
{\tt\small \{yanyichao, whitneypanye, zhuwenhan823\}@sjtu.edu.cn } \\ 
{\tt\small wuwenyan0503@gmail.com daibo@pjlab.org.cn xkyang@sjtu.edu.cn}
}

\let\oldtwocolumn\twocolumn
\renewcommand\twocolumn[1][]{%
	\oldtwocolumn[{#1}{
		\begin{center}
			\includegraphics[width=0.95\linewidth]{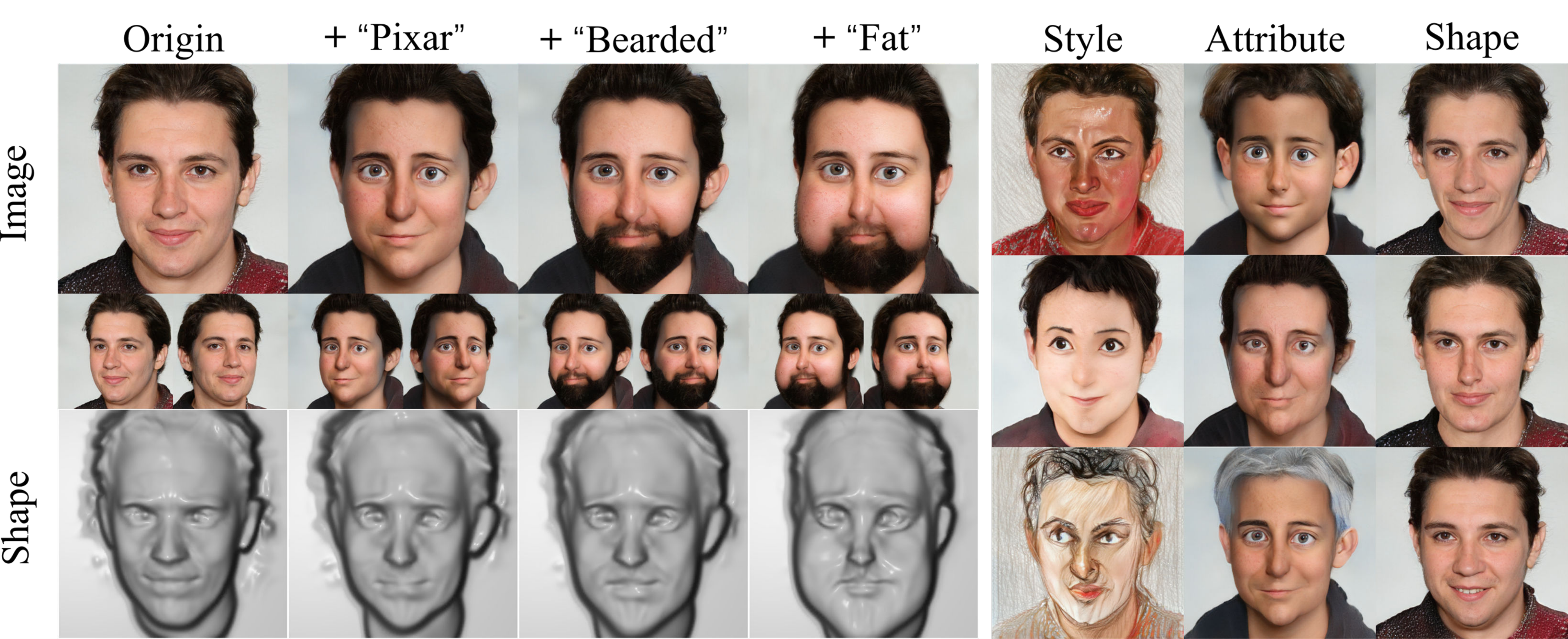}
                \vspace{-0.2 cm}
			\captionof{figure}{Examples of text-guided \textbf{3D portrait stylization}. Our model enables style transfer, attribute editing, shape deformation, and their overlying manipulations under the guidance of text prompts, while preserving 3D consistency.}       
			\label{fig:teaser}
		\end{center}
	}]
}

\maketitle
\input{sections/abs.tex}
\input{sections/intro.tex}
\input{sections/relwork.tex}

\input{sections/method.tex}
\input{sections/exp.tex}

\input{sections/concls.tex}

\ificcvfinal\thispagestyle{empty}\fi

{\small
\bibliographystyle{ieee_fullname}
\bibliography{main}
}

\end{document}

%% file: math_commands/math_commands.tex
\usepackage{amsmath,amsfonts,bm}

\def\eqref#1{equation~\ref{#1}}

\def\1{\bm{1}}

\def\rvz{{\mathbf{z}}}

\def\rmI{{\mathbf{I}}}

\DeclareMathAlphabet{\mathsfit}{\encodingdefault}{\sfdefault}{m}{sl}
\SetMathAlphabet{\mathsfit}{bold}{\encodingdefault}{\sfdefault}{bx}{n}



%% file: sections/abs.tex

\begin{abstract}
Portrait stylization is a long-standing task enabling extensive applications.
Although 2D-based methods have made great progress in recent years, 
real-world applications such as metaverse and games often demand 3D content.
On the other hand,
the requirement of 3D data, which is costly to acquire, significantly impedes the development of 3D portrait stylization methods.
In this paper,
inspired by the success of 3D-aware GANs that bridge 2D and 3D domains with 3D fields as the intermediate representation for rendering 2D images,
we propose a novel method, dubbed HyperStyle3D, based on 3D-aware GANs for 3D portrait stylization.
At the core of our method is a hyper-network learned to manipulate the parameters of the generator in a single forward pass.
It not only offers a strong capacity to handle multiple styles with a single model,
but also enables flexible fine-grained stylization that affects only texture, shape, or local part of the portrait. 
While the use of 3D-aware GANs bypasses the requirement of 3D data,
we further alleviate the necessity of style images with the CLIP model being the stylization guidance.
We conduct an extensive set of experiments across the style, attribute, and shape, and meanwhile, measure the 3D consistency.
These experiments demonstrate the superior capability of our HyperStyle3D model in rendering 3D-consistent images in diverse styles, deforming the face shape, and editing various attributes. 

\end{abstract}

%% file: sections/intro.tex

\section{Introduction}
\label{sec:intro}

Portrait stylization, pervasive in the entertainment industry, aims at transferring the photorealistic face into a target style while preserving the original identity. 
Automatic strategies of neural style transfer, which transfer the given artistic style to a photo based on deep neural networks, have demonstrated the capability of bridging the texture gap between two domains. The impressive effect of style transfer gives rise to a wide range of potential prospects, \ie, virtual makeup, 3D cartoon creation, and short-form video.

Comprehensively considering the application scenarios, an ideal method for portrait stylization is supposed to satisfy three key orthogonal characters, (i) \textbf{natural appearance on style}, (ii) \textbf{flexible deformation on 3D shape}, and (iii) \textbf{light reliance on data.}
In other words, high-quality appearance is the fundamental requirement of style transfer, while the ability of shape deformation contributes to a wider range of applications related to 3D creation, and the light reliance on data can lower the threshold for model training.


Currently, 2D-based methods~\cite{gatys2016image, johnson2016perceptual, ulyanov2017improved, li2017universal, li2018closed, huang2017arbitrary, zhu2017unpaired, patashnik2021styleclip, gal2022stylegan} are capable of rendering the image in the target style while preserving the source content, but portrait stylization is more than just matching the texture style.
Despite the remarkable appearance after stylization, these models fall short of shape deformation and multi-view consistency. 
3D stylization~\cite{han2021exemplar, jung2022deep, lennon2021image2lego} directly operates on the mesh or the voxel, showing the ability to deform shape and compelling 3D consistency on the rendered multi-view images.
However, these methods need dense 3D data of the target style, which demands plenty of effort from professional artists, 
making such data extremely rare and expensive. 
Moreover, these 3D methods are single-style, limited by the specific network and 3D data, and thus incapable of adapting to diverse styles in real-world applications.


Inspired by the recent attempts at stylization in 3D-aware GANs~\cite{zhou2021cips, sun2022ide}, we propose to capitalize on the generative radiance fields for zero-shot portrait stylization, by predicting the parameter offsets of generator driven by the text prompt.
Based on the powerful implicit 3D representations like NeRF~\cite{or2022stylesdf,xu20223d, zhang2022training} and tri-planes~\cite{chan2022efficient}, these 3D-aware GAN models could synthesize high-resolution and 3D consistent images by learning from unposed single-view 2D images only and simultaneously infer the corresponding high-quality 3D shapes, providing a potential model for us to achieve all the three characters of ideal portrait stylization.
With a similar training pipeline as 2D GANs, 3D-aware GANs can leverage the mature style transfer techniques in the 2D domain for appearance. 
More importantly, these methods are also empowered to edit the underlying 3D shapes while bypassing the demands of 3D data.
Meanwhile, in the 2D domain, CLIP-based~\cite{radford2021learning} methods open a new venue for the text-driven manipulation of StyleGAN~\cite{patashnik2021styleclip,gal2022stylegan}, and thereby enable the 3D-aware model to realize zero-shot stylization.
However, employing 2D methods on 3D-aware GANs still faces several challenges.
The optimization of generator parameters~\cite{gal2022stylegan} or the utilization of latent mapper~\cite{patashnik2021styleclip} for a particular input text prompt leads to (i) \textbf{single-style limitation} and (ii) \textbf{incapacity of overlying manipulations}.


To further address the challenges above, we propose HyperStyle3D, an efficient architecture that leverages the hyper-network to bridge the CLIP model and 3D-aware GANs for high-quality zero-shot portrait stylization.
Our hyper-network directly predicts the offset of parameters, getting rid of the time-consuming optimization phase.
The learning capacity of the hyper-network enables it to embed diverse manipulations simultaneously,
and hence we integrate multiple styles in a unified hyper-networks, solving the problem of weak generalization confronted by such single-style methods.
Furthermore, the flexibility of the hyper-network gives rise to another benefit we can explicitly separate different layers of the hyper-network and the generator into multiple levels.
We study the properties of the facial semantics learned by  different layers of the generator and find that face shape, attributes, and appearance style are orthogonally controlled by three different levels of the generator.
Therefore, we split the layers into three groups, \ie, coarse, medium, and fine, whose parameters offset are predicted by three corresponding layer groups of our hyper-network, responsible to shape deformation, attribute editing, and general style transfer, respectively.
Based on this, our hyper-network-based framework bears an additional advantage of overlying manipulations driven by texts with multi-level semantics.

 

Thanks to the proposed hyper-network, our zero-shot 3D-aware stylization model, dubbed HyperStyle3D, overcomes the challenges during the utilization of 3D-aware GANs and successfully meets all three aforementioned characters of an ideal portrait stylization.  
In summary, the contribution of this paper is threefold:
\begin{itemize}
\setlength\itemsep{0em}
\item We introduce the CLIP model to 3D-aware GANs, realizing the zero-shot 3D portrait stylization while avoiding the dependence on expensive 3D data.
\item A novel hyper-network is proposed for multi-style portrait stylization, which enables style transfer, attribute manipulation, shape deformation, and their overlying manipulations in a unified framework.
\item We demonstrate high-quality results on 3D portrait stylization in terms of appearance, shape and 3D consistency as shown in~\cref{fig:teaser}.

\end{itemize}

%% file: sections/relwork.tex

\section{Related Work}
\label{sec:relwork}

\noindent\textbf{2D Style Transfer.}
2D style transfer is an image processing method that renders the semantic content of the image with different styles. A wide variety of image transformation tasks~\cite{gatys2016image, johnson2016perceptual, ulyanov2017improved, li2017universal, li2018closed, huang2017arbitrary} has been proposed based on Convolution Neural Networks (CNN). 
Gatys~\etal~\cite{gatys2016image} first introduces the CNN to the task of style transfer and successfully splits the features into content and style by structuring the GRAM matrix. 
For fast inference during arbitrary style transfer, AdaIN~\cite{huang2017arbitrary} performs style transformations by switching  means and variances of the feature.
With the popularity of 2D GANs, the task of style transfer gradually spreads to unsupervised generative models. 
CycleGAN~\cite{zhu2017unpaired} design a pair of GANs, capable to use the unpaired training data for domain adaptations.
With the advent of StyleGAN~\cite{karras2019style, karras2020analyzing, Karras2021}, plentiful works~\cite{pinkney2020resolution, richardson2021encoding, ojha2021few, jiang2021deceive, huang2021unsupervised, patashnik2021styleclip, gal2022stylegan} propose to synthesize high-quality style images based on this hierarchical style-adaptive framework. 
Recently, to relieve the requirement of style images as training data, StyleCLIP~\cite{patashnik2021styleclip} proposes to use a text to discover global directions in the latent space.
Rather than exploring the latent space, StyleGAN-Nada~\cite{gal2022stylegan} fine-tunes the generator for large changes in the style, beyond the generator’s original domain.
In this work, we extend the style transfer of 2D GANs to 3D-aware GANs, enabling 3D shape deformation and great 3D consistency.

\noindent\textbf{Latent Manipulation.}
Many works explore the latent space of a pre-trained generator for image manipulation~\cite{shen2020interpreting, shen2020interfacegan, shen2021closed, abdal2021styleflow, zhuang2021enjoy, jiang2021talk, voynov2020unsupervised}.
These approaches can be roughly classified into two categories, i.e., 1) unsupervised methods that explore the semantics of generator to discover distinguishable directions~\cite{shen2021closed,voynov2020unsupervised, harkonen2020ganspace} and 2) Supervised methods that use attribute labels to find meaningful latent path~\cite{shen2020interfacegan, shen2020interpreting, zhuang2021enjoy, abdal2021styleflow}.
Sefa~\cite{shen2021closed} proposes a closed-form factorization algorithm to discover latent semantic directions by directly decomposing the pre-trained weights.
StyleFlow~\cite{abdal2021styleflow} utilizes conditional normalizing flow to disentangle the attributes and broaden the editable attributes.
Overall, the manipulation in latent space has demonstrated the ability to perform precise editing of specific attributes.
However, it's not easy to discover the disentangled path in latent space for arbitrary manipulation.
Our HyperStyle3D utilizes hyper-networks to explore the editability in the parameter space of the generator for overlying manipulation of arbitrary styles, attributes, and shapes.

\noindent\textbf{3D Stylization.}
Editing 3D content according to a given style is a challenging task~\cite{xu2020deep,jung2022deep, yin20213dstylenet}, involving both geometric deformation and texture transformation. 
3DStyleNet~\cite{yin20213dstylenet} achieves both shape deformation and texture stylization by a part-aware affine transformation field for shape and a multi-view differentiable renderer for texture editing.
Yang \etal~\cite{jung2022deep} utilizes detailed 3D mesh data and 2D image data of caricature style to train a deformable 3D caricature framework. 
With the advent of NeRF~\cite{martin2020nerf} and the CLIP model~\cite{radford2021learning}, researchers have already made remarkable progress on 3D text-driven synthesis~\cite{wang2022clip,sanghi2022clip, jetchev2021clipmatrix, jain2022zero, michel2022text2mesh}. 
These methods adapt the optimization procedures supervised by the CLIP model~\cite{radford2021learning}. 
Specifically, CLIP-NeRF~\cite{wang2022clip} proposes a unified framework to manipulate NeRF, guided by a text prompt or an example image. 
Given a 3D mesh, Text2Mesh~\cite{michel2022text2mesh} can modify the color and geometry under the guidance of a text prompt.
However, these text-driven methods are all per-instance stylizations that cannot generalize to all objects belonging to a category.
In contrast, our HyperStyle3D is a generative model that edits the face style via learned parameters, capable to apply the manipulation to all the faces.

\noindent\textbf{3D-Aware Image Synthesis.}
Inspired by the superiority of NeRF~\cite{mildenhall2020nerf}~representation,
Several attempts~\cite{schwarz2020graf, chan2020pi, Niemeyer2020GIRAFFE, xu2021generative, pan2021shadegan, niemeyer2021campari, rematasICML21, kosiorek2021nerf, devries2021unconstrained,gu2021stylenerf,zhou2021cips,or2022stylesdf, chan2022efficient, xu20223d, deng2022gram, xiang2022gram,skorokhodov2022epigraf} deploy radiance fields into generative models and thus enable 3D consistent image synthesis.
Recently, 3D-aware GANs~\cite{gu2021stylenerf, zhou2021cips, or2022stylesdf, chan2022efficient, xu20223d, xiang2022gram} combine the shallow NeRF features with a 2D CNN-based renderer and could hallucinate images at $1024^2$ resolution. 
In particular, StyleSDF~\cite{or2022stylesdf} takes Signed Distance Fields (SDF) as the intermediate representation for high-quality shape generation and introduces the regularization loss to preserve 3D consistency.
Nonetheless, expressive 3D-aware GANs have a huge demand for image data and computational sources, arduous to re-train a generator according to the desirable styles.
In our work, we leverage a text-driven zero-shot method to adapt the 3D-aware GANs across various domains, reducing the dependency on style data.

%% file: sections/method.tex
\section{Method}

\begin{figure*}[t]
  \centering
  \begin{subfigure}{\linewidth}
    \includegraphics[width=\linewidth]{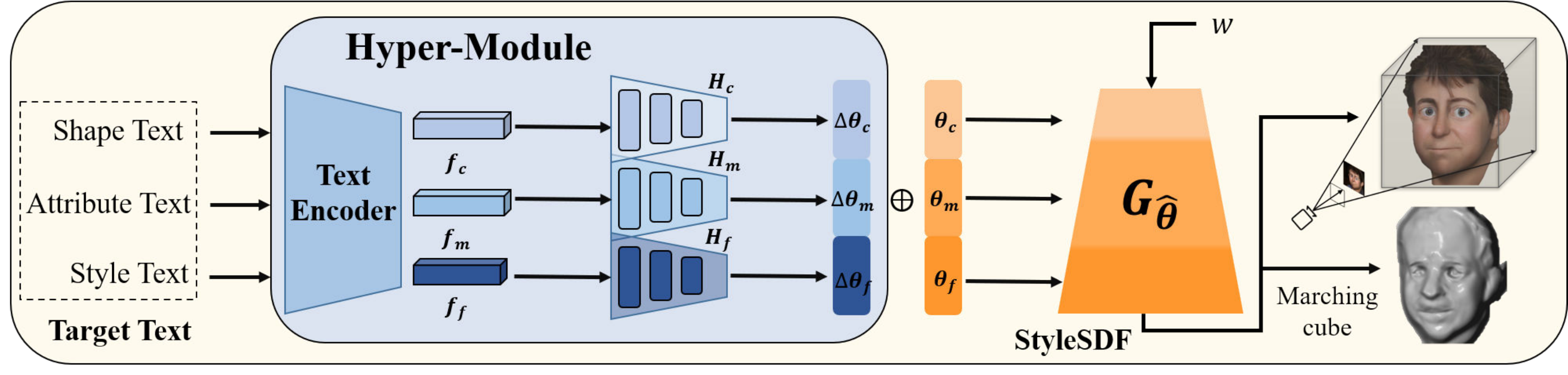}
    \vspace{-0.7cm}
    \caption{3D portrait stylization with \textbf{the hyper-module.} }
    \label{fig:partial-a}
  \end{subfigure}
  
  \centering
  \begin{subfigure}{\linewidth}
    \includegraphics[width=\linewidth]{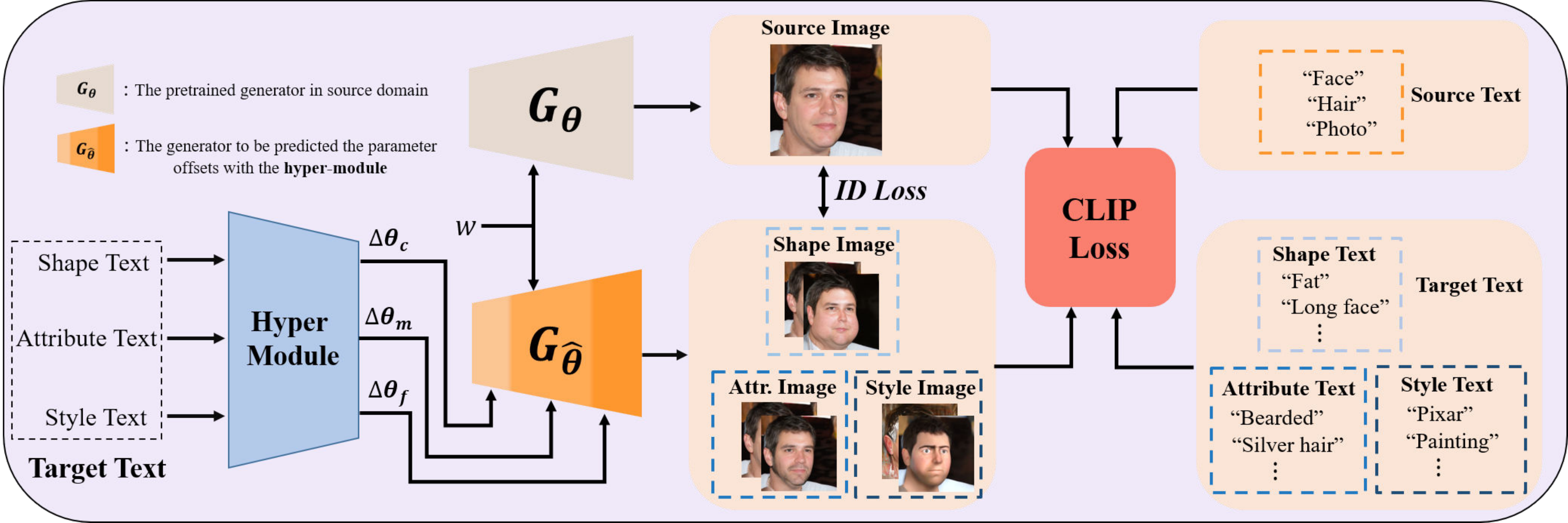}
    \caption{\textbf{Training Pipeline} with multiple styles integrating into a hyper-module.}
    \label{fig:partial-b}
  \end{subfigure}
   \caption{\textbf{The overview} of our full-pipeline. (a) Our hyper-module consists of three trainable hyper-networks and a fixed text encoder. The text prompts of three levels (shape, attribute, and style) are encoded into the coarse, medium, and fine direction features, which are then fed into the corresponding hyper-network. The hyper-networks predict three groups of parameter offsets for the coarse, medium, and fine layers in the pre-trained 3D-aware generator. (b) Our hyper-module is trained under the supervision of CLIP loss and ID loss. Text prompts of three levels are simultaneously integrated into the training, empowering the hyper-module to handle the overlying manipulation of diverse styles, attributes, and shapes. We pre-define the source text as a description related to the current training target text, such as ``Face'' to ``Bearded face''.}
   \vspace{-0.3cm}
   \label{fig:method}
\end{figure*}

Our work aims to enable a pre-trained 3D-aware generator with multi-level manipulations including style transfer, attribute editing, and shape deformation.
For a specific manipulation, our zero-shot method leverages a hyper-network to predict the parameter offsets of the pre-trained generator, under the guidance of corresponding text prompts.
For multi-level manipulations, we split the layers of our hyper-network into multiple groups, each of which corresponds to a specific-level manipulation.
The overview of our architecture is shown in~\cref{fig:method}.

\subsection{Preliminaries on Generative Radiance Fields}\label{subsection:Baseline}

Adopting NeRF~\cite{mildenhall2020nerf} as the 3D representation for 3D-aware GANs has been explored in several works~\cite{schwarz2020graf, chan2020pi, Niemeyer2020GIRAFFE, xu2021generative, pan2021shadegan, niemeyer2021campari, rematasICML21, kosiorek2021nerf, devries2021unconstrained,gu2021stylenerf,zhou2021cips,or2022stylesdf, chan2022efficient, xu20223d, deng2022gram, xiang2022gram,skorokhodov2022epigraf}.
Here we follow the design of StyleSDF~\cite{or2022stylesdf}, one of these awesome models, as the pre-trained generator of the source domain.
StyleSDF is one of the typical NeRF-based GAN models, which adopts the implicit function parameterized as an MLP to represent the 3D scene.
This function takes a 3D coordinate $\textbf{x}  = \left( {x,y,z} \right) \in \mathbb{R}{^3}$ and the view direction $\textbf{d} \in \mathbb{S}{^2}$ as inputs and outputs per-point signed distance values $\textbf{$\alpha$} \left( {x} \right) \in \mathbb{R}{^+}$ and view-dependent color $\textbf{c} \left( \textbf{x},\textbf{d} \right) \in \mathbb{R}{^3}$. 

Unlike pure-MLP generator, instead of directly computing each pixel color for image generation, StyleSDF additionally computes a low-resolution feature map $\textbf{f} \left( \textbf{x},\textbf{d} \right)$ via volume rendering along its corresponding camera ray.
To achieve high-resolution generation under the constraint of memory, it utilizes a similar up-sampling architecture as StyleGAN to efficiently transformed the low-resolution feature maps into high-resolution images.
We summarize this image generation process as $\rmI_g = G_{\theta} (\rvz, \bm{\xi})$, where $G_{\theta}$ is the generator, $\theta$ denotes its learnable parameters, $\rvz$ is the latent code conditioning the generator via mapping networks and $\bm{\xi}$ represents the sampled camera pose.

\subsection{Single-forward Transfer via Hyper-Module}

Previous GAN-based face methods usually perform image manipulation 
in the $\mathcal{W}$ or $\mathcal{W+}$ spaces, however, it is non-trivial to find the specific latent path for the subtle attributions like eyes size without labeled data. It is also difficult to disentangle the latent path, especially for multi-level manipulation.
Additionally, as different identities correspond to different $w$ codes, the semantic direction in latent space is usually instance-specific.

In contrast, parameter space contains more essential and disentangled facial semantics learned by the generator.
Thus, we explore the editability in the parameter space of GANs to discover a general, detailed and disentangled editing path with the guidance of text prompts.
Unlike other methods using per-style optimization process~\cite{gal2022stylegan,patashnik2021styleclip} to fine-tune the parameters, we instead design a single-forward framework based on the hyper-network to predict the parameter offsets of the generator $G_{\theta}$ to simultaneously support multiple styles. 
The refining process is guided by the changing direction of the source text and the target text.

Specifically, we first use a shared text encoder $E$ to transform source text $t_{\rm src}$ and target prompts $t_{\rm tgt}$ into the intermediate features, and then represent the manipulation direction as the feature difference:
\begin{align}
f_{\rm dir} = f_{\rm tgt} - f_{\rm src} = E\left( t_{\rm tgt} \right) - E\left( t_{\rm src} \right).
\label{eq:gan}
\end{align}
As shown in~\cref{fig:method}, the shape text, attribute text, and style text are three classes of target texts. Source texts are the prompts corresponding to each target text, for example, “hair” (source) to “silver hair” (target).

To embed the direction code $f_{\rm dir}$ to the generated image $\rmI_g$, we opt to use $f_{\rm dir}$ to update the pretrained generator $G_{\theta}$.
The updating process is bridged via a series of hyper-networks $H$.
The hyper-network $H$ takes the direction code $f_{\rm dir}$ as input and predicts the parameter offsets $\Delta\theta$, which are then multiplied as a coefficient to refine the parameters of the primary generator $G_{\theta}$.
The offset generator $G_{\hat{\theta}}$ with updated parameters $\hat{\theta}$ enables the generation of manipulated images as the target text describes.
Here we only consider predicting the parameters for the main linear layers of $G_{\theta}$, skipping up-sampling layers, as most of the content is completely generated in the main linear layers.

Assume that the pre-trained generator $G_{\theta}$ has $N$ linear layers with parameters $\theta = \left(\theta_{1}; \theta_{2}; ...; \theta_{N}\right)$. 
As pointed out by the theoretical study of previous works related to the StyleGAN, different layers of the generator contribute to different levels of attributes.
Therefore, we propose to use each small hyper-network $h_{j}$ to predict the parameter offsets $\Delta\theta_{j}$ of each linear layer $j$. 
All the small hyper-networks constitute the integrated hyper-network:
\begin{align}
H = \left\{ h_{j} \right\}, ~j\in\left\{ {1,2,\ldots,N} \right\}.
\label{eq:gan}
\end{align}

To reduce the risk of over-fitting and enhance the diversity of styles, we additionally add some random noise $n^i$ to each style change direction $f^i_{\rm dir}$.
Thus, the parameter offsets $\Delta\theta_{j}$ of the linear layer $j$ is predicted as
\begin{align}
\Delta\theta_{j} = h_{j}\left(f^i_{\rm dir} + n^i\right),~j\in\left\{ {1,2,\ldots,N} \right\}.
\label{eq:gan}
\end{align}
The details of our hyper-network are illustrated in \cref{fig:method}. 
Finally, the updated generator has the parameter as
\begin{align}
\begin{split}
\hat{\theta} = \left\{ {\theta_{j} \cdot \left( {1 + \Delta\theta_{j}} \right)} \right\},~j\in\left\{ {1,2,\ldots,N} \right\}.
\end{split}
\label{eq:gan}
\end{align}
The manipulated image $\rmI_g$ could be generated from the updated generator:
\begin{align}
\rmI_g = G_{\hat{\theta}} (\rvz, \bm{\xi}).
\label{eq:gan}
\end{align}

\subsection{Multi-level Manipulation} \label{subsection:Manipulation}

As a 3D-lifting architecture based on StyleGAN~\cite{karras2019style, karras2020analyzing, Karras2021}, StyleSDF~\cite{or2022stylesdf} displays a similar characteristic that layers make different semantic contributions to the final image generation, and meanwhile this property extends to the 3D geometric contributions due to the 3D representation of the generator. 
Inspired by the Latent Mapper of StyleCLIP~\cite{patashnik2021styleclip}, we investigate the effect of different layers and split them into three groups, \ie, coarse, medium, and fine, according to their contributions,
\ie, shape, attribute, and style, respectively.

Thereby, we extend the hyper-network to also contain three parts along with the three groups of layers. Here, we denote the input text feature as $f~ = ~\left( f_{c};~f_{m};~f_{f} \right)$, corresponding to the three groups of layers. The hyper-network is defined by
\begin{align}
H(f) = \left( {H_{C}\left( f_{c} \right),H_{m}\left( f_{m} \right),H_{f}\left( f_{f} \right)} \right).
\label{aa}
\end{align}

Each group of hyper-network responsible to a type of manipulation (shape, attributes, or styles) is trained to predict the parameter offsets, as indicated by the text prompts of corresponding types, while preserving the other visual attributes of the input image.
\begin{equation}
    {\Delta\theta}_{c} = H_{c}\left( f_{c} \right),
    {\Delta\theta}_{m} = H_{m}\left( f_{m} \right),
    {\Delta\theta}_{f} = H_{f}\left( f_{f} \right).
\end{equation}

Specifically, as the coarse layers are responsible to the fundamental shape generation, the coarse hyper-network $H_{c}$ takes several texts related to the geometric deformation such as “fat” as the training target, and learns to search the corresponding path for shape transformation in parameter space by the supervision of CLIP loss.
Similarly, the medium layers contribute to the facial topology, and thus the medium hyper-network $H_{m}$ is trained for attribute editing. Besides, the fine hyper-network $H_{f}$ is trained for general style transfer because the fine layers mainly influence the color appearance of image results.

The updated parameters are given by:
\begin{align}
    \hat{\theta} = \left( {\theta_{c} + {\alpha_{c}\Delta\theta}_{c},\theta_{m} + {\alpha_{m}\Delta\theta}_{m},\theta_{f} + \alpha_{f}{\Delta\theta}_{f}} \right),
    \label{bb}
\end{align}
where $\alpha_*$ are coefficients indicating the editing degree. 
As $\alpha$ increases, the synthesized images change more positively along the corresponding manipulation. For example, with text “bearded”, the resulting images show a thicker beard as $\alpha$ increases. 
A negative $\alpha$ guides the results to change in the opposite direction of the text description. 
The degree coefficients provide more flexibility for editing.


\begin{figure*}[t]
  \centering
  \includegraphics[width=0.95\linewidth]{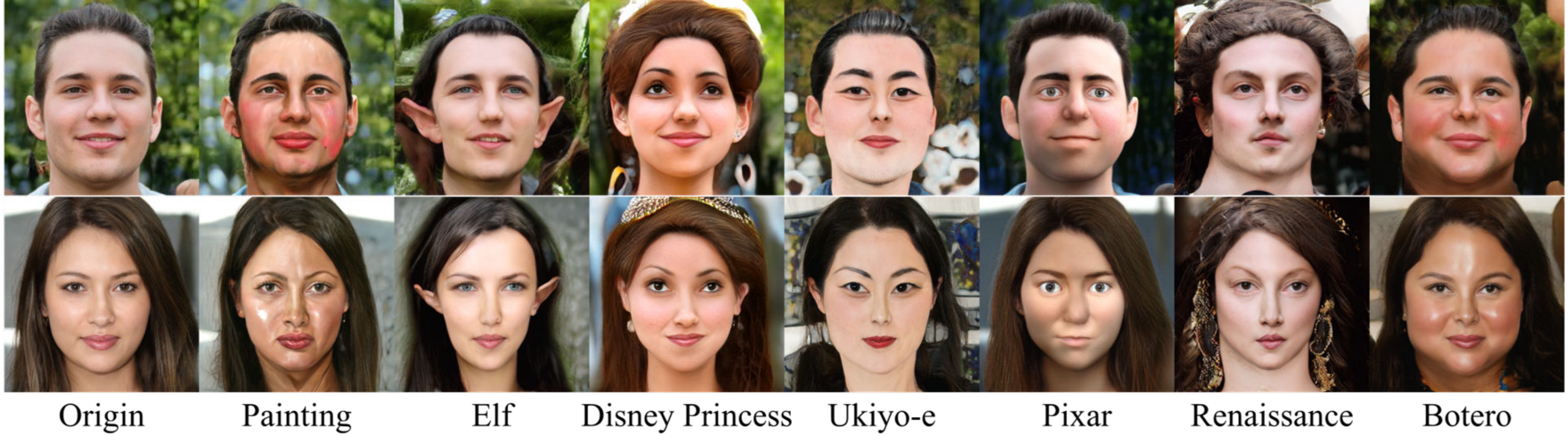}
  \vspace{-0.3cm}
   \caption{\textbf{Qualitative results}. Our hyper-network-based model can transfer the image into diverse style domains.}
   \label{fig:quality_comp}
   \vspace{-0.3cm}
\end{figure*}

\subsection{Training} \label{subsection:Hierarchical Sampling}
\noindent\textbf{CLIP Loss.}
Similar to StyleGAN-Nada~\cite{gal2022stylegan}, our method to predict the parameter offsets of a generator is mainly guided by a text-image directional objective. The direction loss is given by:
\begin{align}
   \mathrm{\Delta}T &= E_{T}\left( t_{\rm tgt} \right) - E_{T}\left( t_{\rm src} \right), \\
   \mathrm{\Delta}I & = E_{I}\left( {G_{\hat{\theta}}(w)} \right) - E_{I}\left( {G_{\theta}}(w) \right), \\
   L_{\rm dir} &= 1 - {\left\langle {\mathrm{\Delta}T,~\mathrm{\Delta}I} 
   \right\rangle}.
   \label{eq:gan}
\end{align}
where $E_{I}$ and $E_{T}$ are the image and text encoders of the CLIP model, $t_{\rm src}$ is the source class text, $t_{\rm tgt}$ is the input text that indicates the target extrinsic style or intrinsic attribution, and $\left\langle \bm{\cdot} \right\rangle$ is the function of cosine similarity. 
The idea is to guide the image produced by the updated generator to change only along the indicated text direction. 

\noindent\textbf{ID Loss.}
For intrinsic attributes manipulation, we further leverage an ID loss~\cite{deng2019arcface} to preserve the facial identity when predicting the parameter offsets of a generator along the text direction. 
Besides, we utilize the generated multi-view images to better keep face identity in 3D space.
The ID loss $L_{\rm ID}$ is given by:
\begin{align}
\begin{split}
L_{\rm ID} = \frac{1}{N} \sum\limits_{i}^{N} \Big\lbrack {1 - \left\langle {F\left( {G_{\hat{\theta}}\left( {w,\xi_{i}} \right)} \right),~F\left( {G_{{\theta}}\left( {w,\xi_{i}} \right)} \right)} \right\rangle} \Big\rbrack,
\label{eq:gan}
\end{split}
\end{align}
where $F\left( \bm{\cdot} \right)$ is a pre-trained ArcFace~\cite{deng2019arcface} model for face recognition to extract identity features, and $i$ indicates the $i$-th view direction of the total $N$ views.
ID loss encourages the extracted features of source and target images to be as close as possible during editing.

\noindent\textbf{Region Loss.}
Although the text description only contains the attributions we want to edit, there exist risks that other irrelevant attributions may change as the generator's parameters shift. To address this, we additionally introduce an optional region loss $L_{\rm region}$ to reduce the risk of entangled attributes in parameters space:
\begin{align}
L_{\rm region} = \left\| {R\left( {G_{{\hat{\theta}}}\left( {w,\xi_{i}} \right)} \right) - R\left( {G_{{\theta}}\left( {w,\xi_{i}} \right)} \right)} \right\|_{2},
\label{eq:gan}
\end{align}
where $R\left( \bm{\cdot} \right)$ expresses the pixels of irrelevant regions predicted by a pre-trained semantic model BiSenet V2~\cite{yu2021bisenet}. The CLIP model is responsible to match the input text to the region. When training the hyper-network for editing attributes, the pixels of irrelevant parts are supervised by the region loss to keep them unchanged as much as possible.

To summarize, the total loss is given by:
\begin{align}
L_{total} = {\lambda_{\rm dir}L}_{\rm dir} + {\lambda_{\rm ID}L}_{\rm ID} + \lambda_{\rm region}L_{\rm region},
\label{eq:gan}
\end{align}
where $\lambda_{\rm direction}$, $\lambda_{\rm ID}$ and $\lambda_{\rm region}$ are the weights for CLIP loss, ID loss, and region loss, respectively.

%% file: sections/exp.tex


\section{Experiments}

\subsection{Qualitative results}

\begin{figure}[t]
  \centering
  \includegraphics[width=\linewidth]{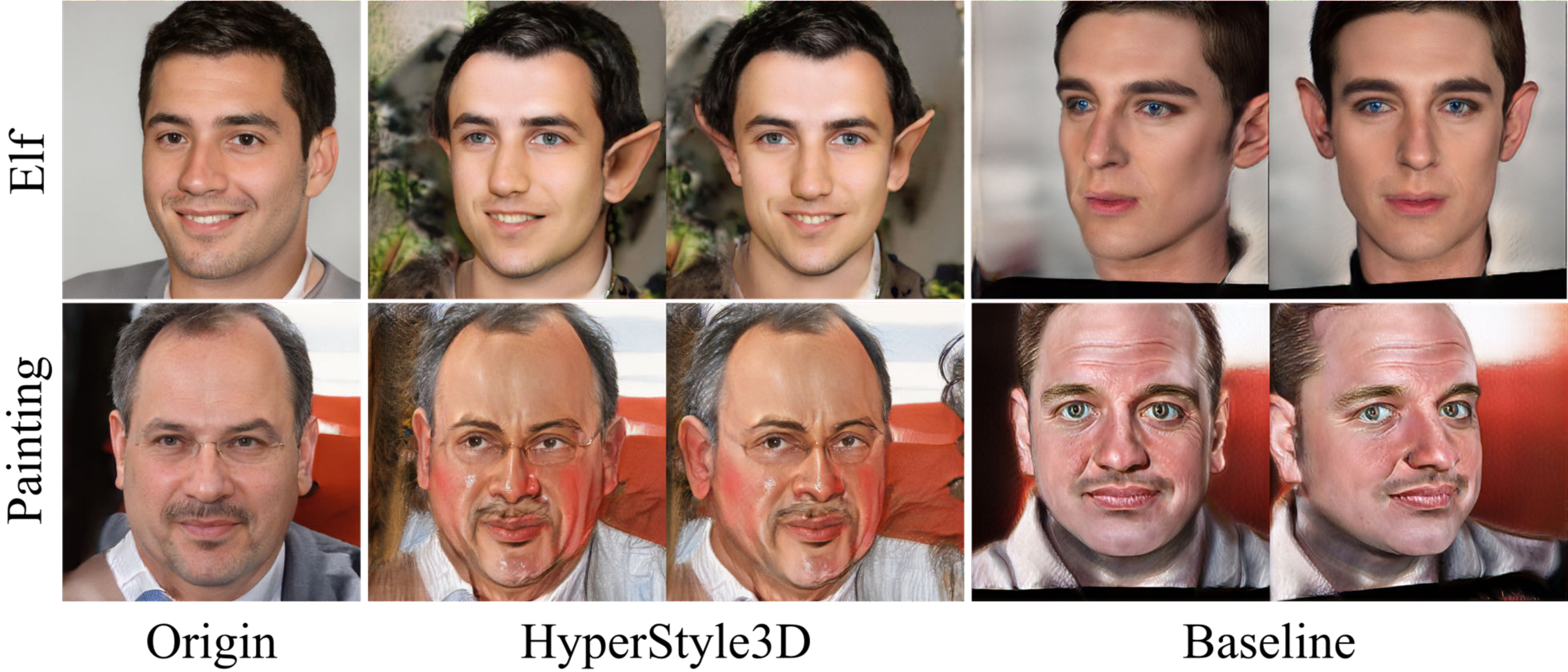}
  \vspace{-0.7cm}
   \caption{\textbf{Qualitative comparison}. Our HyperStyle3D achieves better style realism and recovers more face details compared to the baseline, \ie, 2D transfer + 3D GAN inversion. It's noteworthy that 3D GAN inversion has to operate on the novel style domain, resulting in the loss of rich face details accordingly.}
   \label{fig:Qualitative_comp}
   \vspace{-0.5cm}
\end{figure}

In this section, we discuss the experiments conducted to evaluate the quality of stylized images.
Our generator starts from the real-image domain of the FFHQ dataset~\cite{karras2019style}, to the multiple out-of-domain styles, \ie, ``Painting'', ``Elf'', ``Disney Princess'', ``Ukiyo-e'', ``Pixar'', ``Renaissance'' and ``Botero''.
As shown in~\cref{fig:quality_comp}, our hyper-network-based model can synthesize high-quality stylized images.
Although all the styles are embedded in a unified hyper-network, it can be noticed that our model still adapts to a wide range of styles beyond the pre-trained generator's domain.
Moreover, as shown in~\cref{fig:Qualitative_comp}, we compare our HyperStyle3D with the baseline that first adopts 2D style transfer techniques~\cite{rombach2022high} on the images and then inverts stylized images to the latent space of a pre-trained 3D-aware GAN~\cite{chan2022efficient}. We additionally perform a user study in~\cref{tab:user}.
Both qualitatively and quantitatively, our model demonstrates better performance with regard to style realism and face details.

\begin{table}[h]
	\centering
	\vspace{-0.2cm}
        \resizebox{0.9\linewidth}{!}{
	\begin{tabular}{@{}lcc@{}}
		\toprule
		Method & Style realism $\uparrow$ & Face details $\uparrow$  \\
		\midrule
		Baseline~\cite{chan2022efficient,rombach2022high} & 2.8  & 3.4 \\
		Ours & \textbf{4.0}  & \textbf{3.8} \\
		\bottomrule
	\end{tabular}}
	\vspace{-0.2cm}
	\caption{\textbf{User Study} conducted with 20 participants. The range of scores is from 1 to 5. The higher, the better. 
    }
	\label{tab:user}
	\vspace{-0.5cm}
\end{table}

\subsection{Quantitative results}
One of the advances of 3D stylization compared to the 2D methods is the 3D consistency of generated images.
To show that our methods have not undermined the 3D consistency of pre-trained 3D-aware generators, we conduct a quantitative experiment that adopts the depth consistency used in StyleSDF~\cite{or2022stylesdf} and facial identity consistency used in EG3D~\cite{chan2022efficient} as the evaluation metrics of 3D consistency.


\begin{table}
  \centering
  \begin{tabular}{@{}lcc@{}}
    \toprule
    Method & Depth error $\downarrow$ & ID similarity $\uparrow$  \\
    \midrule
    EG3D~\cite{chan2022efficient}            & 4.47  & 0.80 \\
    Original StyleSDF~\cite{or2022stylesdf}  & 1.48  & 0.81 \\
    Hyper-networks for shape                & 1.52  & 0.84 \\
    Shape + attribute                       & 1.52  & 0.88 \\
    Shape + attribute + style               & 1.55  & -    \\
    \bottomrule
  \end{tabular}
  \vspace{-0.3cm}
  \caption{\textbf{Quantitative results of 3D consistency.} Our model is based on StyleSDF and shows comparable performance after the stylization. Even with overlying manipulations, our model will not be degraded in terms of depth consistency and ID consistency between various views. Arcface~\cite{deng2019arcface} pre-trained model cannot be adapted to style domains, and thus the ID similarity result of stylized images is not provided here.}
  \vspace{-0.5cm}
  \label{tab:3D_consistency}
\end{table}

\noindent\textbf{Depth Consistency.}
To evaluate the depth consistency, we sample 500 identities, render their $128 \times 128$ depth maps from the frontal view and a random side view, and compute the depth error after the alignment between the two views.
A modified Chamfer distance metric is adopted as the measurement of alignment error between two depth points.
As shown in~\cref{tab:3D_consistency}, despite stylization and multiple manipulations, our hyper-networks have achieved comparable depth consistency compared to the original StyleSDF~\cite{or2022stylesdf}.

\noindent\textbf{Facial Identity Consistency.}
The ArcFace~\cite{deng2019arcface} cosine similarity of facial identity between multi-views is a common metric for the evaluation of 3D consistency adopted in 3D-aware GANs~\cite{chan2022efficient,zhang2022training}.
To evaluate facial identity consistency, we sample 1500 identities, render their $512 \times 512$ high-resolution images from two random side views, and measure the cosine similarity of those two views.
As the ArcFace model is pre-trained on the FFHQ dataset, it is not able to evaluate the identity similarity of style images.
Therefore, facial identity consistency is only used for the evaluation of attribute editing and shape deformation.
As shown in~\cref{tab:3D_consistency}, after shape deformation and attribute editing, portrait images generated by ours are even superior to those from the original model in terms of ID similarity, demonstrating the preservation of ID consistency of our method.
The superior result is because the exaggerated shape and attribute make the facial features more distinctive.

\begin{figure}[t]
  \centering
  \includegraphics[width=0.95\linewidth]{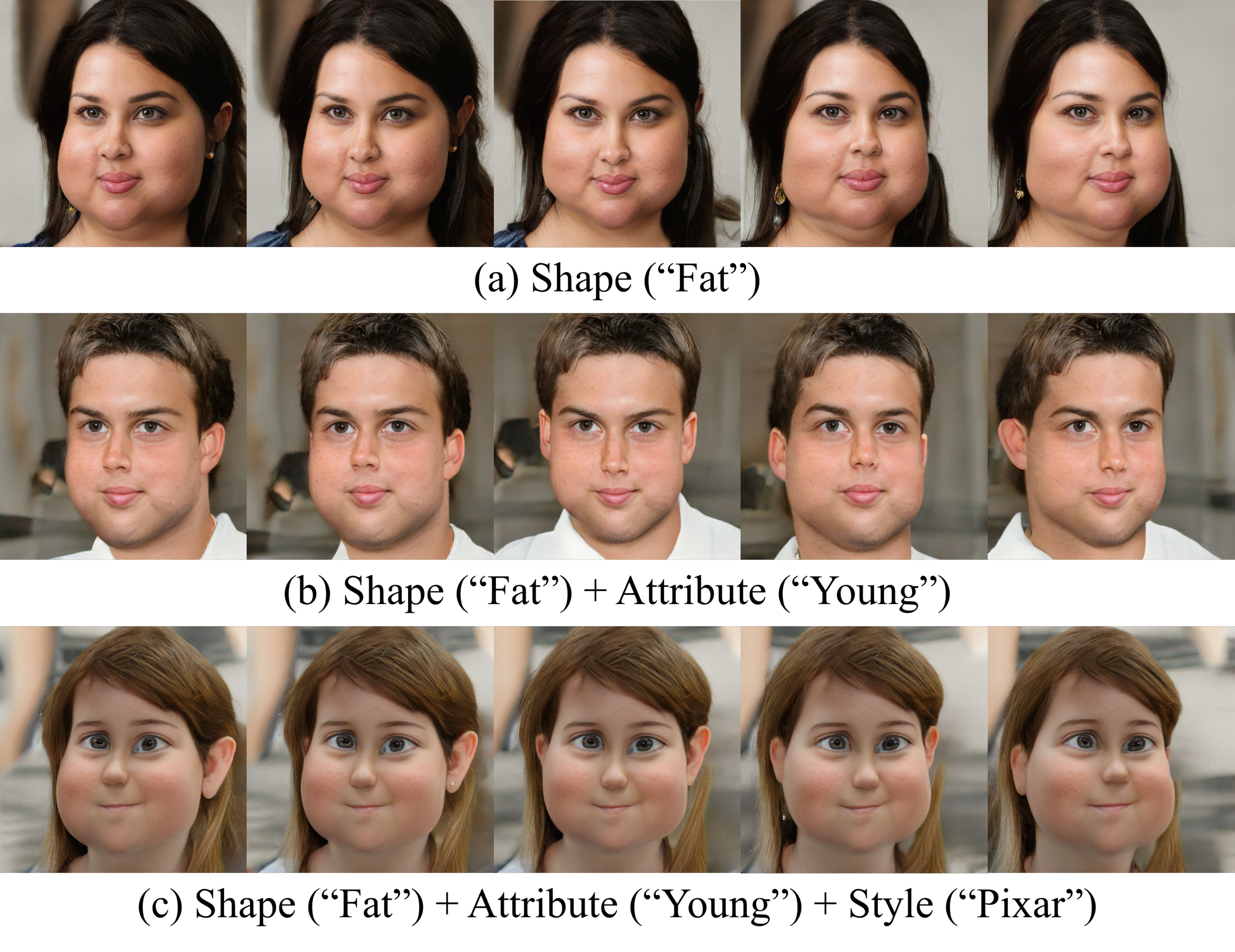}
  \vspace{-0.3cm}
   \caption{\textbf{The results of 3D consistency}. We  sample several view directions to show the 3D consistency. As can be observed, the 3D consistency is still well maintained after the manipulation via hyper-network.}
   \vspace{-0.5cm}
   \label{fig:consistency}
\end{figure}

We also show several examples with multiple views in~\cref{fig:consistency} for the illustration of 3D consistency, while additional cases with more views are provided in the \emph{supplementary material}.

\subsection{Component Analysis}

\begin{figure}[t]
  \centering
  \includegraphics[width=\linewidth]{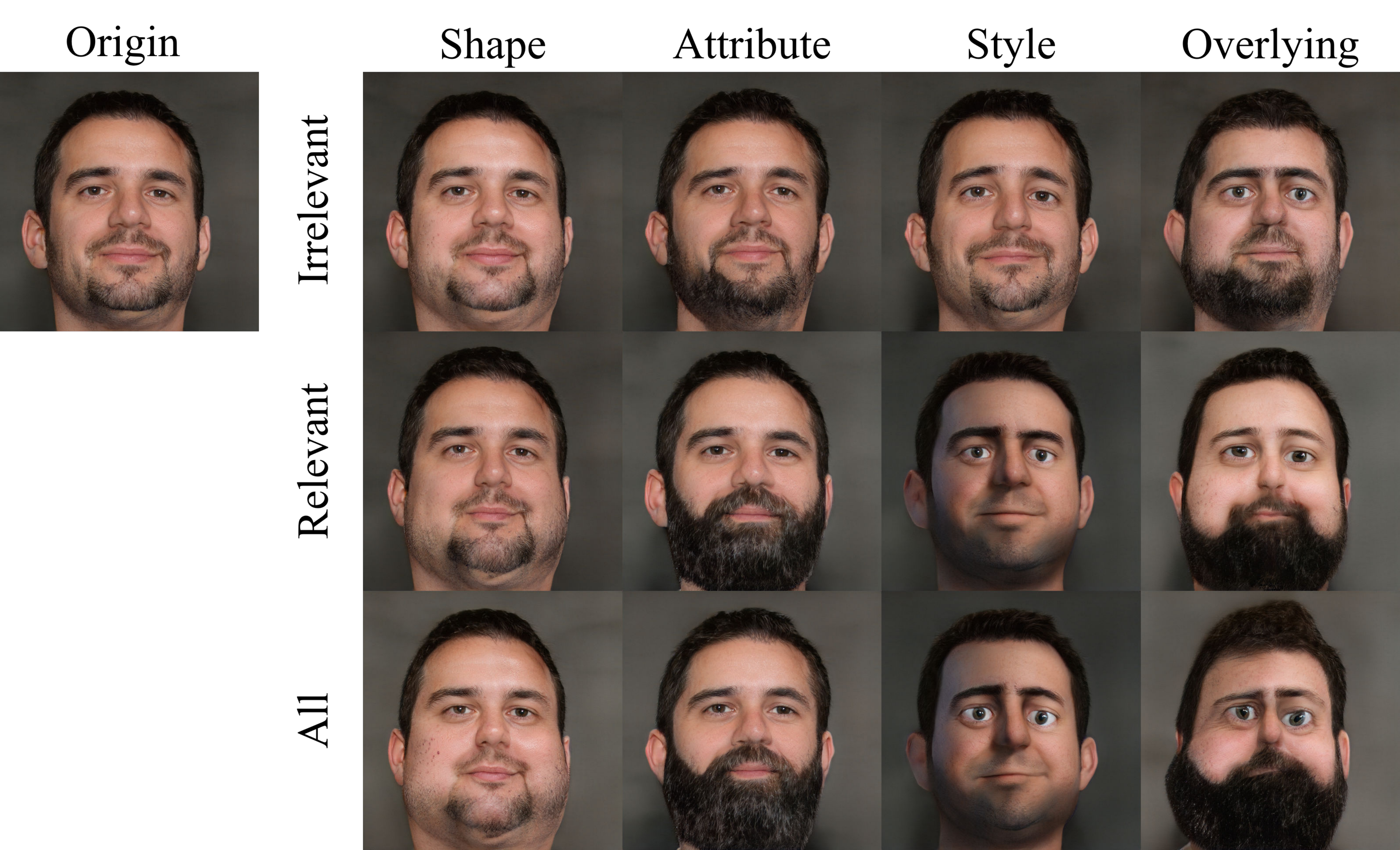}
  \vspace{-0.7cm}
   \caption{\textbf{Ablation study} of the effect of different layers. For a specific level text prompt, parameter offsets on the relevant (the same level) layers lead to a significant change, otherwise, the change brought by the parameter update on irrelevant layers is less notable. Blindly updating parameters of all layers can work in the single style mode, but results in degenerate images in the overlying manipulation.}
   \vspace{-0.5cm}
   \label{fig:ablation_diff}
\end{figure}

\begin{figure*}[t]
  \centering
  \begin{subfigure}{\linewidth}
    \includegraphics[width=0.95\linewidth]{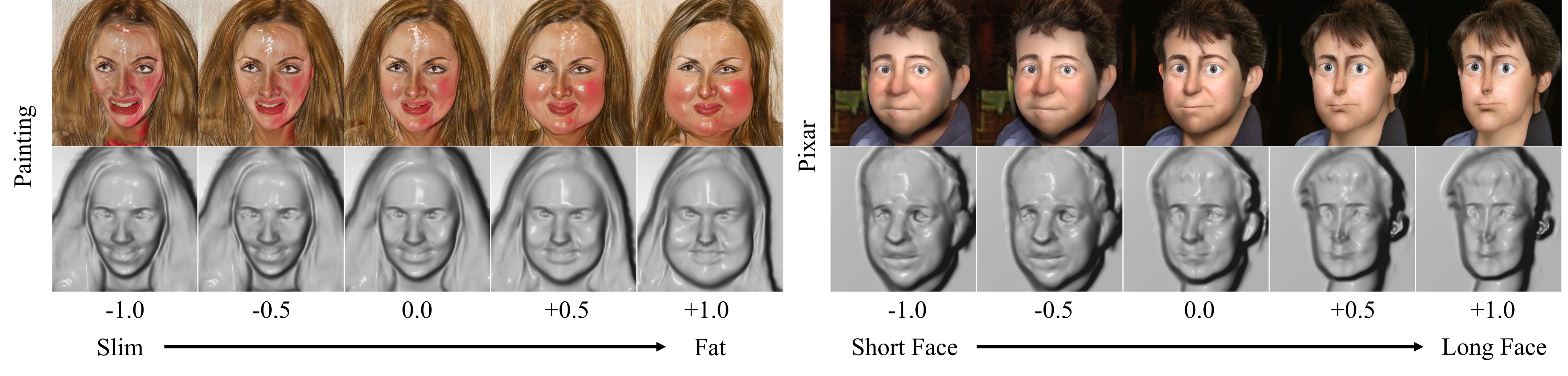}
    \vspace{-0.3cm}
    \caption{\textbf{Shape deformation} under the control of coefficient.}
    \label{fig:shape_coe}
  \end{subfigure}
  
  \centering
  \begin{subfigure}{\linewidth}
    \includegraphics[width=0.97\linewidth]{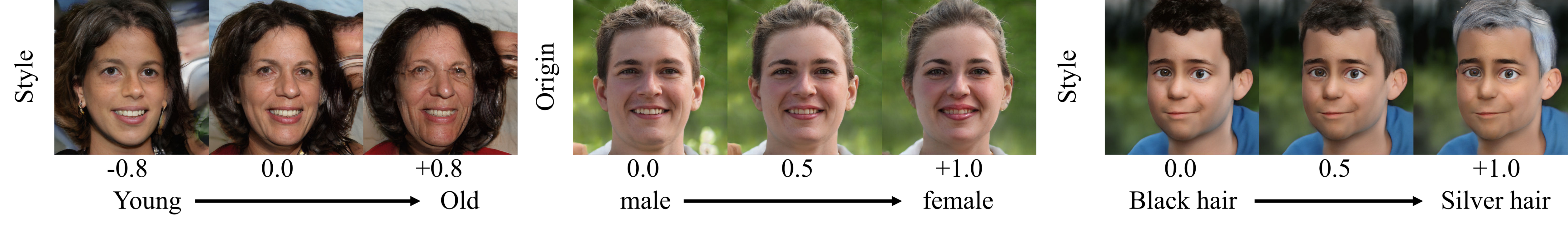}
    \vspace{-0.3cm}
    \caption{\textbf{Attribute editing} under the control of coefficient.}
    \label{fig:ablation_attr}
  \end{subfigure}
  \vspace{-0.65cm}
   \caption{\textbf{Results of controllable degree of manipulations}. As the coefficient $\alpha$ increases, the manipulated image \textbf{gradually} changes towards the target direction. With a negative coefficient, the image varies along the opposite direction of the target text. As we can see, shape and attribute show a similar trend.}
   \vspace{-0.5cm}
   \label{fig:coefficient}
\end{figure*}

\begin{figure}[h]
	\centering
	\includegraphics[width=\linewidth]{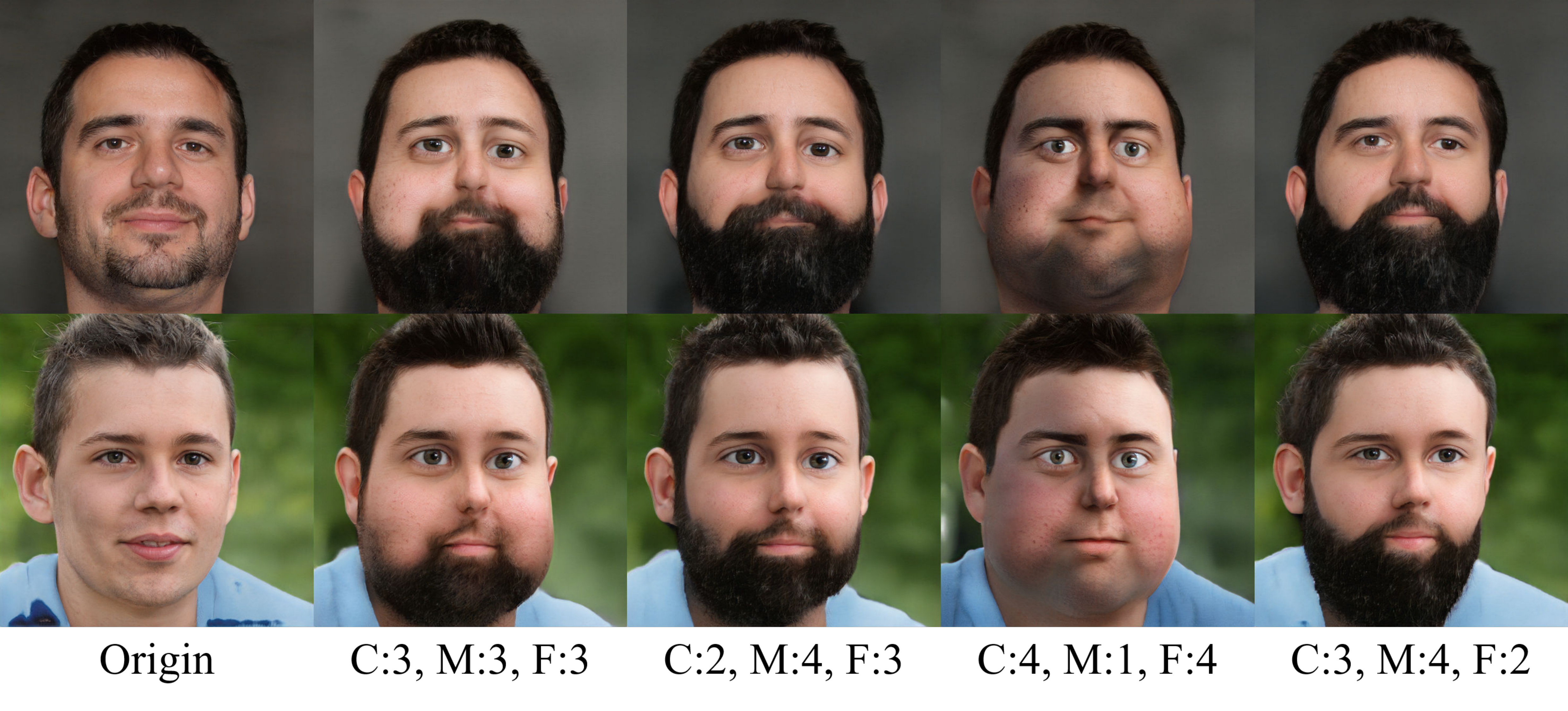}
	\vspace{-0.8cm}
	\caption{\textbf{Ablation study} on different choices of layer groups. For example, ``C:2, M:4, F:3'' means that the first two layers serve as the coarse group responsible to shape deformation, the middle four layers are the medium group responsible to attribute manipulation, and the last three layers serve as the fine group responsible to style transfer.}
	\vspace{-0.4cm}
	\label{fig:ablation_layer}
\end{figure}

\noindent\textbf{Multi-level Manipulation.}
Due to the different levels of manipulations, it's beneficial to analyze the layer mechanism and accordingly split hyper-networks into shape-controller, attribute controller, and style-controller.
Aiming to achieve disentangled manipulations of multiple levels, we explore the effect of each layer in StyleSDF~\cite{or2022stylesdf} contributing to image generation. 
In~\cref{fig:ablation_diff}, we compare the performance of several models with different updated layers according to the different training text prompts.
To ensure a fair comparison, all models are trained with the same batch size and learning rate.

As illustrated in~\cref{fig:ablation_diff}, with the text prompt “fat”, predicting the parameter offset of the coarse layers leads to significant shape deformation, but updating others is relatively useless.
For the text prompts of attribute and style, it also shows a similar trend that fine-tuning medium and fine layers can lead to a significant change in attribute and style, respectively, otherwise, the change brought by the parameter update on irrelevant layers is less notable.
Besides, as shown in the last column, with all layers to update, the generator feels confused when each layer simultaneously receives three different parameter offsets predicted by hyper-networks, resulting in erratic generated images.

Hence, we can conclude that the shape is mainly controlled by the coarse layers, while the medium and fine layers contribute little to the shape deformation.
Similarly, the attribute and the style are dominantly influenced by the medium layers and the fine layers, respectively.

\noindent\textbf{Controllable Degree of Manipulation.}
As described in~\cref{subsection:Manipulation}, we leverage the hyper-network to predict the parameter offsets for shape deformation, attribution manipulation, and style transfer. Furthermore, the degree of attribute and shape can be linearly controlled by the interpolations done in parameter space. 
\cref{fig:shape_coe} visualizes the controllability aspect of the shape deformation process based on a coefficient $\alpha$ multiplied by the parameter offsets.
Taking ``Fat'' as an example, with the coefficient rising from $0$ to $1$, the generated face gradually turns chubby, while a decreasing coefficient can also slim the face. 
A similar trend can be also found with other shape prompts such as ``Long face''.

\cref{fig:ablation_attr} visualizes results of attribute editing controlled by the coefficient $\alpha$.
With the target text prompt of ``Old'', ``Female'', and ``Silver hair'', it is shown that the woman turns old, the male becomes female, and the hair color becomes silver, respectively, as the coefficient $\alpha$ increases.

\noindent\textbf{Ablation of group choices.}
Based on the study of multi-level manipulation, we further verify the impact of different choices of layer groups, \ie, coarse, medium, and fine.
To this end, we conduct an additional ablation study in \cref{fig:ablation_layer}, with a combination of text prompts, \ie, ``Fat'', ``Bearded Face'', and ``Pixar''. The main part of StyleSDF~\cite{or2022stylesdf} contains nine linear layers. As can be observed, the division of C: 3, M: 3, and F: 3 can balance all three aspects, \ie, shape, attribute, and style, while other divisions tend to lack capacity in one respect. For example, the division of C: 2, M: 4, and F: 3 cannot adapt to large shape deformation, the division of C: 4, M: 1, and F: 4 fails to edit an attribute, and the division of C: 3, M: 4, and F: 2 shows little change of style.
Hence, we set the coarse/medium/fine layers to 3/3/3 respectively, so as to balance the manipulation of shape, attributes, and style.

%% file: sections/concls.tex
\section{Conclusion}

In this work, we propose HyperStyle3D, an efficient text-driven method based on 3D-aware GANs for 3D portrait stylization.
Specifically, we introduce a hyper-network to predict the offset of the generator parameters with the guidance of the CLIP model.
As shown in the experiments, our model achieves high-quality results with regard to style transfer, attribute editing, and shape deformation while avoiding the reliance on rare 3D data.
Moreover, we find the style, attribute, and shape are controlled by three separated layer groups in the hyper-network, based on which multi-level overlying manipulations can be realized.

%% file: main.bbl
\begin{thebibliography}{10}\itemsep=-1pt

\bibitem{abdal2021styleflow}
Rameen Abdal, Peihao Zhu, Niloy~J Mitra, and Peter Wonka.
\newblock Styleflow: Attribute-conditioned exploration of stylegan-generated
  images using conditional continuous normalizing flows.
\newblock {\em ACM Transactions on Graphics (ToG)}, pages 1--21, 2021.

\bibitem{chan2022efficient}
Eric~R Chan, Connor~Z Lin, Matthew~A Chan, Koki Nagano, Boxiao Pan, Shalini
  De~Mello, Orazio Gallo, Leonidas~J Guibas, Jonathan Tremblay, Sameh Khamis,
  et~al.
\newblock Efficient geometry-aware 3d generative adversarial networks.
\newblock In {\em CVPR}, pages 16123--16133, 2022.

\bibitem{chan2020pi}
Eric~R Chan, Marco Monteiro, Petr Kellnhofer, Jiajun Wu, and Gordon Wetzstein.
\newblock pi-gan: Periodic implicit generative adversarial networks for
  3d-aware image synthesis.
\newblock In {\em CVPR}, pages 5799--5809, 2021.

\bibitem{deng2019arcface}
Jiankang Deng, Jia Guo, Niannan Xue, and Stefanos Zafeiriou.
\newblock Arcface: Additive angular margin loss for deep face recognition.
\newblock In {\em CVPR}, pages 4690--4699, 2019.

\bibitem{deng2022gram}
Yu Deng, Jiaolong Yang, Jianfeng Xiang, and Xin Tong.
\newblock Gram: Generative radiance manifolds for 3d-aware image generation.
\newblock In {\em CVPR}, pages 10673--10683, 2022.

\bibitem{devries2021unconstrained}
Terrance DeVries, Miguel~Angel Bautista, Nitish Srivastava, Graham~W Taylor,
  and Joshua~M Susskind.
\newblock Unconstrained scene generation with locally conditioned radiance
  fields.
\newblock In {\em CVPR}, pages 14304--14313, 2021.

\bibitem{gal2022stylegan}
Rinon Gal, Or Patashnik, Haggai Maron, Amit~H Bermano, Gal Chechik, and Daniel
  Cohen-Or.
\newblock Stylegan-nada: Clip-guided domain adaptation of image generators.
\newblock {\em ACM Transactions on Graphics (TOG)}, pages 1--13, 2022.

\bibitem{gatys2016image}
Leon~A Gatys, Alexander~S Ecker, and Matthias Bethge.
\newblock Image style transfer using convolutional neural networks.
\newblock In {\em CVPR}, pages 2414--2423, 2016.

\bibitem{gu2021stylenerf}
Jiatao Gu, Lingjie Liu, Peng Wang, and Christian Theobalt.
\newblock Stylenerf: A style-based 3d aware generator for high-resolution image
  synthesis.
\newblock In {\em ICLR}, 2021.

\bibitem{han2021exemplar}
Fangzhou Han, Shuquan Ye, Mingming He, Menglei Chai, and Jing Liao.
\newblock Exemplar-based 3d portrait stylization.
\newblock {\em IEEE Transactions on Visualization and Computer Graphics}, 2021.

\bibitem{harkonen2020ganspace}
Erik H{\"a}rk{\"o}nen, Aaron Hertzmann, Jaakko Lehtinen, and Sylvain Paris.
\newblock Ganspace: Discovering interpretable gan controls.
\newblock {\em arXiv preprint arXiv:2004.02546}, 2020.

\bibitem{huang2021unsupervised}
Jialu Huang, Jing Liao, and Sam Kwong.
\newblock Unsupervised image-to-image translation via pre-trained stylegan2
  network.
\newblock {\em IEEE Transactions on Multimedia}, pages 1435--1448, 2021.

\bibitem{huang2017arbitrary}
Xun Huang and Serge Belongie.
\newblock Arbitrary style transfer in real-time with adaptive instance
  normalization.
\newblock In {\em ICCV}, pages 1501--1510, 2017.

\bibitem{jain2022zero}
Ajay Jain, Ben Mildenhall, Jonathan~T Barron, Pieter Abbeel, and Ben Poole.
\newblock Zero-shot text-guided object generation with dream fields.
\newblock In {\em CVPR}, pages 867--876, 2022.

\bibitem{jetchev2021clipmatrix}
Nikolay Jetchev.
\newblock Clipmatrix: Text-controlled creation of 3d textured meshes.
\newblock {\em arXiv preprint arXiv:2109.12922}, 2021.

\bibitem{jiang2021deceive}
Liming Jiang, Bo Dai, Wayne Wu, and Chen~Change Loy.
\newblock Deceive d: Adaptive pseudo augmentation for gan training with limited
  data.
\newblock In {\em NIPS}, pages 21655--21667, 2021.

\bibitem{jiang2021talk}
Yuming Jiang, Ziqi Huang, Xingang Pan, Chen~Change Loy, and Ziwei Liu.
\newblock Talk-to-edit: Fine-grained facial editing via dialog.
\newblock In {\em ICCV}, pages 13799--13808, 2021.

\bibitem{johnson2016perceptual}
Justin Johnson, Alexandre Alahi, and Li Fei-Fei.
\newblock Perceptual losses for real-time style transfer and super-resolution.
\newblock In {\em ECCV}, pages 694--711, 2016.

\bibitem{jung2022deep}
Yucheol Jung, Wonjong Jang, Soongjin Kim, Jiaolong Yang, Xin Tong, and
  Seungyong Lee.
\newblock Deep deformable 3d caricatures with learned shape control.
\newblock In {\em ACM SIGGRAPH 2022 Conference Proceedings}, pages 1--9, 2022.

\bibitem{Karras2021}
Tero Karras, Miika Aittala, Samuli Laine, Erik H\"ark\"onen, Janne Hellsten,
  Jaakko Lehtinen, and Timo Aila.
\newblock Alias-free generative adversarial networks.
\newblock In {\em NIPS}, pages 852--863, 2021.

\bibitem{karras2019style}
Tero Karras, Samuli Laine, and Timo Aila.
\newblock A style-based generator architecture for generative adversarial
  networks.
\newblock In {\em CVPR}, pages 4401--4410, 2019.

\bibitem{karras2020analyzing}
Tero Karras, Samuli Laine, Miika Aittala, Janne Hellsten, Jaakko Lehtinen, and
  Timo Aila.
\newblock Analyzing and improving the image quality of stylegan.
\newblock In {\em CVPR}, pages 8110--8119, 2020.

\bibitem{kosiorek2021nerf}
Adam~R Kosiorek, Heiko Strathmann, Daniel Zoran, Pol Moreno, Rosalia Schneider,
  So{\v{n}}a Mokr{\'a}, and Danilo~J Rezende.
\newblock Nerf-vae: A geometry aware 3d scene generative model.
\newblock {\em arXiv preprint arXiv:2104.00587}, 2021.

\bibitem{lennon2021image2lego}
Kyle Lennon, Katharina Fransen, Alexander O'Brien, Yumeng Cao, Matthew
  Beveridge, Yamin Arefeen, Nikhil Singh, and Iddo Drori.
\newblock Image2lego: Customized lego set generation from images.
\newblock {\em arXiv preprint arXiv:2108.08477}, 2021.

\bibitem{li2017universal}
Yijun Li, Chen Fang, Jimei Yang, Zhaowen Wang, Xin Lu, and Ming-Hsuan Yang.
\newblock Universal style transfer via feature transforms.
\newblock In {\em NIPS}, 2017.

\bibitem{li2018closed}
Yijun Li, Ming-Yu Liu, Xueting Li, Ming-Hsuan Yang, and Jan Kautz.
\newblock A closed-form solution to photorealistic image stylization.
\newblock In {\em ECCV}, 2018.

\bibitem{martin2020nerf}
Ricardo Martin-Brualla, Noha Radwan, Mehdi~SM Sajjadi, Jonathan~T Barron,
  Alexey Dosovitskiy, and Daniel Duckworth.
\newblock Nerf in the wild: Neural radiance fields for unconstrained photo
  collections.
\newblock {\em arXiv preprint arXiv:2008.02268}, 2020.

\bibitem{michel2022text2mesh}
Oscar Michel, Roi Bar-On, Richard Liu, Sagie Benaim, and Rana Hanocka.
\newblock Text2mesh: Text-driven neural stylization for meshes.
\newblock In {\em CVPR}, pages 13492--13502, 2022.

\bibitem{mildenhall2020nerf}
Ben Mildenhall, Pratul~P Srinivasan, Matthew Tancik, Jonathan~T Barron, Ravi
  Ramamoorthi, and Ren Ng.
\newblock Nerf: Representing scenes as neural radiance fields for view
  synthesis.
\newblock In {\em ECCV}, pages 99--106, 2020.

\bibitem{niemeyer2021campari}
Michael Niemeyer and Andreas Geiger.
\newblock Campari: Camera-aware decomposed generative neural radiance fields.
\newblock {\em arXiv preprint arXiv:2103.17269}, 2021.

\bibitem{Niemeyer2020GIRAFFE}
Michael Niemeyer and Andreas Geiger.
\newblock Giraffe: Representing scenes as compositional generative neural
  feature fields.
\newblock In {\em CVPR}, pages 11453--11464, 2021.

\bibitem{ojha2021few}
Utkarsh Ojha, Yijun Li, Jingwan Lu, Alexei~A Efros, Yong~Jae Lee, Eli
  Shechtman, and Richard Zhang.
\newblock Few-shot image generation via cross-domain correspondence.
\newblock In {\em CVPR}, pages 10743--10752, 2021.

\bibitem{or2022stylesdf}
Roy Or-El, Xuan Luo, Mengyi Shan, Eli Shechtman, Jeong~Joon Park, and Ira
  Kemelmacher-Shlizerman.
\newblock Stylesdf: High-resolution 3d-consistent image and geometry
  generation.
\newblock In {\em CVPR}, pages 13503--13513, 2022.

\bibitem{pan2021shadegan}
Xingang Pan, Xudong Xu, Chen~Change Loy, Christian Theobalt, and Bo Dai.
\newblock A shading-guided generative implicit model for shape-accurate
  3d-aware image synthesis.
\newblock In {\em NIPS}, 2021.

\bibitem{patashnik2021styleclip}
Or Patashnik, Zongze Wu, Eli Shechtman, Daniel Cohen-Or, and Dani Lischinski.
\newblock Styleclip: Text-driven manipulation of stylegan imagery.
\newblock In {\em CVPR}, pages 2085--2094, 2021.

\bibitem{pinkney2020resolution}
Justin~NM Pinkney and Doron Adler.
\newblock Resolution dependent gan interpolation for controllable image
  synthesis between domains.
\newblock {\em arXiv preprint arXiv:2010.05334}, 2020.

\bibitem{radford2021learning}
Alec Radford, Jong~Wook Kim, Chris Hallacy, Aditya Ramesh, Gabriel Goh,
  Sandhini Agarwal, Girish Sastry, Amanda Askell, Pamela Mishkin, Jack Clark,
  et~al.
\newblock Learning transferable visual models from natural language
  supervision.
\newblock In {\em ICML}, pages 8748--8763, 2021.

\bibitem{rematasICML21}
Konstantinos Rematas, Ricardo Martin-Brualla, and Vittorio Ferrari.
\newblock Sharf: Shape-conditioned radiance fields from a single view.
\newblock In {\em ICML}, 2021.

\bibitem{richardson2021encoding}
Elad Richardson, Yuval Alaluf, Or Patashnik, Yotam Nitzan, Yaniv Azar, Stav
  Shapiro, and Daniel Cohen-Or.
\newblock Encoding in style: a stylegan encoder for image-to-image translation.
\newblock In {\em CVPR}, pages 2287--2296, 2021.

\bibitem{rombach2022high}
Robin Rombach, Andreas Blattmann, Dominik Lorenz, Patrick Esser, and Bj{\"o}rn
  Ommer.
\newblock High-resolution image synthesis with latent diffusion models.
\newblock In {\em Proceedings of the IEEE/CVF Conference on Computer Vision and
  Pattern Recognition}, pages 10684--10695, 2022.

\bibitem{sanghi2022clip}
Aditya Sanghi, Hang Chu, Joseph~G Lambourne, Ye Wang, Chin-Yi Cheng, Marco
  Fumero, and Kamal~Rahimi Malekshan.
\newblock Clip-forge: Towards zero-shot text-to-shape generation.
\newblock In {\em CVPR}, pages 18603--18613, 2022.

\bibitem{schwarz2020graf}
Katja Schwarz, Yiyi Liao, Michael Niemeyer, and Andreas Geiger.
\newblock Graf: Generative radiance fields for 3d-aware image synthesis.
\newblock In {\em NIPS}, 2020.

\bibitem{shen2020interpreting}
Yujun Shen, Jinjin Gu, Xiaoou Tang, and Bolei Zhou.
\newblock Interpreting the latent space of gans for semantic face editing.
\newblock In {\em CVPR}, pages 9243--9252, 2020.

\bibitem{shen2020interfacegan}
Yujun Shen, Ceyuan Yang, Xiaoou Tang, and Bolei Zhou.
\newblock Interfacegan: Interpreting the disentangled face representation
  learned by gans.
\newblock {\em IEEE transactions on pattern analysis and machine intelligence},
  2020.

\bibitem{shen2021closed}
Yujun Shen and Bolei Zhou.
\newblock Closed-form factorization of latent semantics in gans.
\newblock In {\em CVPR}, pages 453--468, 2021.

\bibitem{skorokhodov2022epigraf}
Ivan Skorokhodov, Sergey Tulyakov, Yiqun Wang, and Peter Wonka.
\newblock Epigraf: Rethinking training of 3d gans.
\newblock {\em arXiv preprint arXiv:2206.10535}, 2022.

\bibitem{sun2022ide}
Jingxiang Sun, Xuan Wang, Yichun Shi, Lizhen Wang, Jue Wang, and Yebin Liu.
\newblock Ide-3d: Interactive disentangled editing for high-resolution 3d-aware
  portrait synthesis.
\newblock {\em arXiv preprint arXiv:2205.15517}, 2022.

\bibitem{ulyanov2017improved}
Dmitry Ulyanov, Andrea Vedaldi, and Victor Lempitsky.
\newblock Improved texture networks: Maximizing quality and diversity in
  feed-forward stylization and texture synthesis.
\newblock In {\em CVPR}, pages 6924--6932, 2017.

\bibitem{voynov2020unsupervised}
Andrey Voynov and Artem Babenko.
\newblock Unsupervised discovery of interpretable directions in the gan latent
  space.
\newblock In {\em ICML}, pages 9786--9796, 2020.

\bibitem{wang2022clip}
Can Wang, Menglei Chai, Mingming He, Dongdong Chen, and Jing Liao.
\newblock Clip-nerf: Text-and-image driven manipulation of neural radiance
  fields.
\newblock In {\em CVPR}, pages 3835--3844, 2022.

\bibitem{xiang2022gram}
Jianfeng Xiang, Jiaolong Yang, Yu Deng, and Xin Tong.
\newblock Gram-hd: 3d-consistent image generation at high resolution with
  generative radiance manifolds.
\newblock {\em arXiv preprint arXiv:2206.07255}, 2022.

\bibitem{xu2020deep}
Sicheng Xu, Jiaolong Yang, Dong Chen, Fang Wen, Yu Deng, Yunde Jia, and Xin
  Tong.
\newblock Deep 3d portrait from a single image.
\newblock In {\em CVPR}, pages 7710--7720, 2020.

\bibitem{xu2021generative}
Xudong Xu, Xingang Pan, Dahua Lin, and Bo Dai.
\newblock Generative occupancy fields for 3d surface-aware image synthesis.
\newblock In {\em NIPS}, 2021.

\bibitem{xu20223d}
Yinghao Xu, Sida Peng, Ceyuan Yang, Yujun Shen, and Bolei Zhou.
\newblock 3d-aware image synthesis via learning structural and textural
  representations.
\newblock In {\em CVPR}, 2022.

\bibitem{yin20213dstylenet}
Kangxue Yin, Jun Gao, Maria Shugrina, Sameh Khamis, and Sanja Fidler.
\newblock 3dstylenet: Creating 3d shapes with geometric and texture style
  variations.
\newblock In {\em ICCV}, 2021.

\bibitem{yu2021bisenet}
Changqian Yu, Changxin Gao, Jingbo Wang, Gang Yu, Chunhua Shen, and Nong Sang.
\newblock Bisenet v2: Bilateral network with guided aggregation for real-time
  semantic segmentation.
\newblock {\em International Journal of Computer Vision}, 129:3051--3068, 2021.

\bibitem{zhang2022training}
Jichao Zhang, Aliaksandr Siarohin, Yahui Liu, Hao Tang, Nicu Sebe, and Wei
  Wang.
\newblock Training and tuning generative neural radiance fields for
  attribute-conditional 3d-aware face generation.
\newblock {\em arXiv preprint arXiv:2208.12550}, 2022.

\bibitem{zhou2021cips}
Peng Zhou, Lingxi Xie, Bingbing Ni, and Qi Tian.
\newblock Cips-3d: A 3d-aware generator of gans based on
  conditionally-independent pixel synthesis.
\newblock {\em arXiv preprint arXiv:2110.09788}, 2021.

\bibitem{zhu2017unpaired}
Jun-Yan Zhu, Taesung Park, Phillip Isola, and Alexei~A Efros.
\newblock Unpaired image-to-image translation using cycle-consistent
  adversarial networks.
\newblock In {\em ICCV}, pages 2223--2232, 2017.

\bibitem{zhuang2021enjoy}
Peiye Zhuang, Oluwasanmi Koyejo, and Alexander~G Schwing.
\newblock Enjoy your editing: Controllable gans for image editing via latent
  space navigation.
\newblock {\em arXiv preprint arXiv:2102.01187}, 2021.

\end{thebibliography}
